\documentclass[runningheads]{llncs}
\usepackage{algorithmic}
\usepackage{algorithm}
\usepackage{hyperref}
\usepackage{times}
\usepackage{epsfig}
\usepackage{graphicx}
\usepackage{amsmath}
\usepackage{amssymb}
\usepackage{multirow}
\usepackage{scrextend}
\usepackage[inline]{enumitem}
\usepackage{xcolor}
\usepackage[bold]{hhtensor}
\usepackage[T1]{fontenc}
\usepackage{color}
\usepackage{float}
\usepackage{placeins}
\graphicspath{{figures/}}
\usepackage[font=normal]{subfig}
\usepackage{enumitem}
\usepackage{multirow}
\usepackage{array}
\usepackage{textcomp}
\usepackage[misc]{ifsym}

\makeatletter
\newcommand{\printfnsymbol}[1]{\textsuperscript{\@fnsymbol{#1}}}
\makeatother
\newcommand{\PreserveBackslash}[1]{\let\temp=\\#1\let\\=\temp}
\newcolumntype{C}[1]{>{\PreserveBackslash\centering}p{#1}}

\begin{document}
\title{Comparing to Learn: Surpassing ImageNet Pretraining on Radiographs By Comparing Image Representations}
%
%
\author{Hong-Yu Zhou\thanks{First two authors contributed equally.} \and
Shuang Yu\printfnsymbol{1} \and Cheng Bian \Letter \and
Yifan Hu \and Kai Ma \and Yefeng Zheng} 

%
\authorrunning{}
%
\institute{Jarvis Lab, Tencent\\
\email{\{hongyuzhou, shirlyyu, tronbian, ivanyfhu, kylekma, yefengzheng\}@tencent.com}}
\maketitle              
\begin{abstract}
	In deep learning era, pretrained models play an important role in medical image analysis, in which ImageNet pretraining has been widely adopted as the best way. However, it is undeniable that there exists an obvious domain gap between natural images and medical images. To bridge this gap, we propose a new pretraining method which learns from 700k radiographs given no manual annotations. We call our method as \emph{Comparing to Learn} (C2L) because it learns robust features by comparing different image representations. To verify the effectiveness of C2L, we conduct comprehensive ablation studies and evaluate it on different tasks and datasets. The experimental results on radiographs show that C2L can outperform ImageNet pretraining and previous state-of-the-art approaches significantly. Code and models are available at \url{https://github.com/funnyzhou/C2L_MICCAI2020}.
\end{abstract}
\section{Introduction}
ImageNet~\cite{deng2009imagenet} pretraining has been proved to be an effective way to perform 2D transfer learning for medical image analysis. Lots of experiments have shown that, compared with learning from scratch, pretrained models not only help to achieve higher accuracy but also speed up the model convergence. These benefits can be attributed to two factors: (a) effective learning algorithms designed for deep neural networks and (b) generalized feature representations learned from a great quantity of natural images. However, there exists an obvious domain gap between natural images and medical images, which raises a question whether we can have a pretrained model directly from medical images and how to approach this. 

As is well known, we need domain experts' diagnosis to produce reliable medical annotations, which definitely help improve our model performance. On the other side, it is often difficult to access to a great number of doctors' conclusions considering limited medical resources as well as protection of patient privacy. So how to develop algorithms to learn from a vast amount of data without annotations has drawn more attention in the medical imaging community. Zhou \emph{et al.}~\cite{zhou2019models} proposed a self-supervised pretraining method \emph{Model Genesis} which utilized medical images without manual labeling. On the chest X-ray classification task, Model Genesis is able to achieve comparable performance with ImageNet pretraining but still cannot beat it. 

In this paper, we present a novel self-supervised pretraining approach focusing on providing pretrained 2D deep models for radiograph related tasks from massive unannotated data. We name our method as Comparing to Learn (C2L) because the goal is to learn general image representations by comparing different image features as the supervision. 
Different from Model Genesis~\cite{zhou2019models} that resorts to an image restoration pretext task, the supervision signal of the proposed C2L comes from the self-defined representation similarity. Similar ideas have been adopted in~\cite{caron2018deep,wang2017transitive,he2019momentum}, where most of them take advantage of the transitive invariance of images to produce self-supervised signal. On the contrary, in this paper, we mainly focus on feature level contrast and propose to construct homogeneous and heterogeneous data pairs by mixing image and feature batches. Moreover, a momentum-based teacher-student architecture is proposed for the contrastive learning, where the teacher and student networks share the same structure but are updated differently. To be specific, the teacher model is updated using both itself and the student network. Extensive experiments of different datasets and tasks demonstrate that the proposed C2L method can surpass ImageNet pretraining and other competitive baselines by non-trivial margins. 



\begin{figure}[t]
	\centering
	\includegraphics[width=0.8\columnwidth]{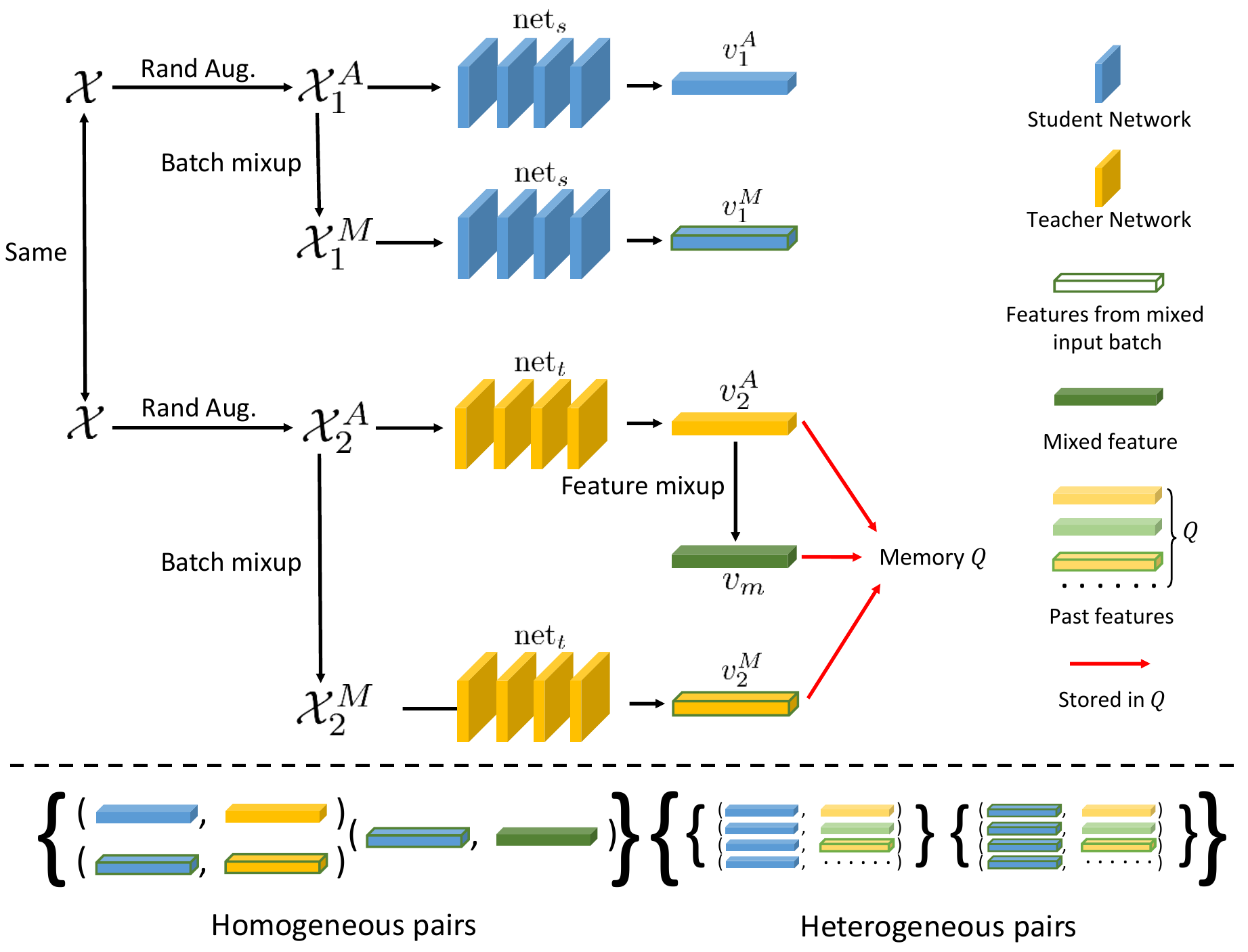}
	\caption{Overview of proposed Comparing to Learn (C2L) framework. For clarify, we also demonstrate our definition of homogeneous and heterogeneous image pairs. Note that light color volumes in $Q$ denote past features for $v_2^A$, $v_m$ and $v_2^M$, respectively.}
	\label{framework}
\end{figure}
\section{Proposed Method}
In this section, we introduce the proposed Comparing to Learn (C2L) method in details. 
The overall workflow is provided in Figure~\ref{framework}.

\begin{algorithm}[t] 
	\caption{The detailed training procedure of C2L.} 
	\label{procedure} 
	\begin{algorithmic}[1]
		\REQUIRE ~~\\
		The original input image batch $\mathcal{X}$ which contains $Z$ images;\\
		\STATE $\mathcal{X}_{1}^{A}=\text{Augment}(\mathcal{X})$; $\mathcal{X}_{2}^{A}=\text{Augment}(\mathcal{X})$ {\color{gray} $/\!\!/$ Generate two augmented batches}
		\STATE $\mathcal{X}_{1}^{M}=\text{mixup}(\mathcal{X}_{1}^{A})$; $\mathcal{X}_{2}^{M}=\text{mixup}(\mathcal{X}_{2}^{A})$ {\color{gray} $/\!\!/$ Apply mixup within each batch}
		\STATE Initialize two networks $\text{net}_s$ and $\text{net}_t$
		\STATE {\color{gray} $/\!\!/$ Q is a memory queue storing past features. Its length is $N$ and $D$ is the feature dimension}
		\STATE $\text{Q}=\text{Rand}(N, D)$
		\STATE \textbf{for} $i = 1$ \textbf{to} $K$ \textbf{do} {\color{gray} $/\!\!/$ $K$ is the number of training iterations}
		\STATE \ \ \ \ \ $\text{v}^A_1=\text{net}_s(\mathcal{X}^A_1)$; $\text{v}^M_1=\text{net}_s(\mathcal{X}^M_1)$ {\color{gray} $/\!\!/$ we use $v$ to stand for output feature vectors}
		\STATE \ \ \ \ \ $\text{v}^A_2=\text{net}_t(\mathcal{X}^A_2)$; $\text{v}^M_2=\text{net}_t(\mathcal{X}^M_2);v_m=\text{mixup}(\text{v}^A_2)$
		\STATE \ \ \ \ \ $\text{loss}_{A} = 0;\text{loss}_{M}=0$
		\STATE \ \ \ \ \ \textbf{for} $j = 1$ \textbf{to} $Z$ \textbf{do}
		\STATE \ \ \ \ \ \ \ \ \ \ $/\!\!/$ {\color{gray} $gt$ stands for the one-hot ground truth vector \{1,0,0,...,0\} whose size is N+1}
		\STATE \ \ \ \ \ \ \ \ \ \ $\text{loss}_{A}$+$=\text{CE}(\left\{(v_{1,j}^A)^Tv_{2,j}^A,(v_{1,j}^A)^TQ\right\},gt)$\label{inner1}
		\STATE \ \ \ \ \ \ \ \ \ \  $\text{loss}_{M}$+$=\text{CE}(\left\{(v_{1,j}^M)^Tv_{2,j}^M,(v_{1,j}^M)^TQ\right\},gt)+\text{CE}(\left\{(v_{1,j}^M)^Tv_{m,j},(v_{1,j}^M)^TQ\right\},gt)$\label{inner2}
		\STATE \ \ \ \ \ \textbf{end for}
		\STATE \ \ \ \ \ $\text{loss} = \text{loss}_{A} + \text{loss}_{M}$
		\STATE \ \ \ \ \ $\text{Backward}\ (\text{loss})$ {\color{gray} $/\!\!/$ Update net$_s$}
		\STATE \ \ \ \ \ $\text{net}_t=\text{Momentum}(\text{net}_t, \text{net}_s)$ {\color{gray} $/\!\!/$ Update net$_t$}
		\label{moment}
		\STATE \ \ \ \ \ $\text{Q}.\text{insert}(v_2^A,v_2^M,v_m)$ {\color{gray} $/\!\!/$ Update $Q$ using current feature vectors.} \label{Q}
		\STATE \textbf{end for}
	\end{algorithmic}
\end{algorithm}

\textbf{Batch mixup and feature mixup.} As shown in Figure~\ref{framework} and Algorithm~\ref{procedure}, for each input image batch, we first use random augmentation (e.g., random cropping, rotation, and cutout~\cite{devries2017improved}) to generate two augmented batches. Different from traditional image-level mixup~\cite{zhang2017mixup}, a batch-wise mixup operation is proposed to apply to each augmented batch. Suppose each batch $\mathcal{X}^A_i$ contains $Z$ images where $i=\left\{1,2\right\}$, we randomly shuffle $\mathcal{X}^A_i$ to construct its paired batch $\mathcal{X}^M_i$ which can be expressed as:
\begin{equation}
\begin{aligned}
\mathcal{X}^M_i =  \lambda*\mathcal{X}^A_i+(1-\lambda)*\text{shuffle}(\mathcal{X}^A_i),
\end{aligned}
\end{equation}
where $\lambda\sim\text{Beta}(1.0,1.0)$ and Beta stands for the beta distribution. In practice, we found that \emph{using the same mixing factor $\lambda$} and shuffling method for both batches ($i=\left\{1,2\right\}$) actually helps improve the model performance. As for feature mixup in Figure~\ref{framework}, we apply the same mixing strategy to the feature representations.

\textbf{Teacher network.} An intuitive idea is to use the same model for both student and teacher networks. However, we found that such strategy does not work in practice which may lead to gradient explosion. Meanwhile, constructing teacher network using momentum update has been widely adopted as a way to produce stable predictions~\cite{tarvainen2017mean,he2019momentum}. In our case, using momentum helps stabilize the training process and reduce the difficulty of network optimization. As shown in line~\ref{moment} from Algorithm~\ref{procedure}, the momentum function can be formalized as:
\begin{equation}
\begin{aligned}
\text{net}_t=\theta*\text{net}_t+(1-\theta)*\text{net}_s,
\end{aligned}
\label{momentum}
\end{equation}
where we use an exponential factor $\theta$ to control the degree of momentum. We can see that the teacher model $\text{net}_t$ is updated using both itself and the student network $\text{net}_s$. In practice, we pass $\mathcal{X}_1^A$ and $\mathcal{X}_1^M$ to the student network while $\mathcal{X}_2^A$ and $\mathcal{X}_2^M$ are passed to the teacher network. In Algorithm~\ref{procedure}, we use $v_1^{\left\{A,M\right\}}$ to represent feature vectors from the student model and $v_2^{\left\{A,M\right\}}$ are used to represent the outputs of the teacher model.

\textbf{Homogeneous and heterogeneous pairs}. For constructing homogeneous pairs, we assume that applying data augmentation (including mixup operation) only slightly change the distribution of training data. Based on this, each homogeneous pair should contain the results after the same set of operations which includes random augmentation, batch mixup or feature mixup as shown in Figure~\ref{framework}. For heterogeneous pairs, we simply contrast the current features with \emph{all} preceding features stored in the memory queue.

\textbf{Feature comparison, memory $Q$ and loss function}. As we have mentioned above, the goal of C2L is to minimize the distance between homogeneous representation pairs such as $(v^A_1, v^A_2)$ and $(v^M_1, v^M_2)$. Meanwhile, it is also necessary to maximize the difference between heterogeneous representations, in which we contrast current features with past features which are collected from past training iterations. To store these past representations, we employ a memory queue $Q$ proposed in \cite{he2019momentum}. The reason why we use a large $Q$ is that we hope to contrast current features with a great number of preceding features because more comparisons usually lead to better representations (as shown in Table~\ref{aug}). So the pairs of current features and past features can be formalized as ($v^A_1$, $Q$) and ($v^M_1$, $Q$), as shown in Figure~\ref{framework}. For simplicity, we use $v^Q_i$ to denote a specific feature vector in $Q$ where $i=\left\{1,2,...,N\right\}$ and $N$ is the length of queue. To be specific, we import a subscript $j$ to index a feature vector given a batch of features. Then, we can convert this distance measurement problem to a naive classification problem. For image $j$ in $\mathcal{X}^A_1$, $\{(v_{1,j}^A)^Tv_{2,j}^A,(v_{1,j}^A)^TQ\}$ in line~\ref{inner1} can be further expressed as:
\begin{align}
\left\{(v^A_{1,j})^T v^A_{2,j},\ (v^A_{1,j})^T v^Q_1,\ (v^A_{1,j})^T v^Q_2,\ ...,\ (v^A_{1,j})^T v^Q_N\right\},
\end{align}
whose length is $N+1$. A similar case also exists in $\mathcal{X}^M_1$. For more comparison, we apply feature mixup to $v^A_2$. In Algorithm~\ref{procedure}, we use $v_m$ to represent the output of feature mixup. Similarly, we also compare $v^M_{1}$ with $v_m$ and $Q$ which leads to another set of $N+1$ predictions:
\begin{align}
\left\{(v^M_{1,j})^T v_{m,j},\ (v^M_{1,j})^T v^Q_1,\ (v^M_{1,j})^T v^Q_2,\ ...,\ (v^M_{1,j})^T v^Q_N\right\}.
\end{align}
It is worth noting that \emph{the first item in each set should be larger than other items because the first item is an inner product of homogeneous representation}. Thus, we can apply a cross entropy loss (CE) to the above sets of predictions where a one hot vector \{1,0,0,...,0\} is treated as the ground truth for each set. 

After we update the network parameters, we also update $Q$ by inserting $v_2^A$, $v_2^M$, and $v_m$, respectively. Since $Q$ is a queue and has a fixed size, the previous feature vectors are automatically removed. After we complete the training stage, only $\text{net}_s$ is extracted to become a pretrained model.

\section{Datasets}
\subsection{Pretraining}
ImageNet pretraining contains about one million labeled images which helps deep models learn general representations. In this paper, we make use of ChestX-ray14~\cite{wang2017chestx}, MIMIC-CXR~\cite{johnson2019mimic}, CheXpert~\cite{irvin2019chexpert} and MURA~\cite{rajpurkar2017mura} as unlabeled data for network pretraining. Note that we only use ChestX-ray14 in ablation studies in order to choose appropriate hyperparameters. After that, we merge four datasets and discard their labels to perform unsupervised pretraining which has approximate 700k unlabeled radiographs.

\textbf{ChestX-ray14.} The training set contains 86k images while the validation set has 25k X-rays. For the ablation study, 70k images from the training set are used for self-supervised pretraining and the rest 16k images are used for fine-tuning to show the results of pretraining. Moreover, after we determine the appropriate hyperparameters, we merge the whole training set into the other three datasets. Overall, C2L uses about 700k unlabeled radiographs for model pretraining.

\textbf{CheXpert.} The training set has 220k images while the official validation set contains 234 images. Similar to ChestX-ray14, we only use the training set without labels for self-supervised pretraining.

\textbf{MIMIC-CXR.} The MIMIC-CXR dataset is a large publicly available dataset of chest radiographs in the JPEG format with structured labels derived from free-text radiology reports. The dataset contains 377,110 JPEG format images. In practice, we treat the whole dataset as an unlabeled database.

\textbf{MURA.} MURA is a dataset of bone X-rays. The training set contains 36k X-rays and the validation set contains 3k images. The whole dataset is used for C2L pretraining.

\subsection{Fine-tuning}
We fine-tune our pretrained models on ChestX-ray14, CheXpert and Kaggle Pneumonia Detection and report their experimental results. 

\textbf{ChestX-ray14 and CheXpert.} For ChestX-ray14, we use all labeled X-rays in the training set (86k) to fine-tune models pretrained with C2L and report experimental results on the validation set. The same setting also applies to CheXpert where we use the whole labeled training set (220k) for fine-tuning.

\textbf{Kaggle Pneumonia Detection.} This dataset is designed for diagnosing pneumonia automatically and accurately. We split the training set in Stage 1 into a local training set (80\%) and a validation set (20\%). The evaluation metric is mean average precision.

\section{Implementation Details}
For pretraining, we employ C2L to pretrain ResNet-18 and DenseNet-121. The default batch size is 256 and the size of each input image is 224$\times$224. For input augmentation, we apply random crop, rotation (10 degree), grayscale and horizontal flip to each input batch. Moreover, we also add cutout to augmented images in order to increase the diversity of transformation for learning better representations. We use L2 normalization for each feature vector.
The momentum factor $\theta$ is set to 0.999 and the length of queue $Q$ is $2^{15}=32768$. We use SGD as the default optimizer where the initial learning rate is 0.03 and its weight decay is 0.0001. We train each model for 240 epochs and the learning rate is divided by 10 at 120, 160 and 200 epochs, respectively.
When fine-tuning pretrained models on ChestX-ray14 and CheXpert, the input image size is set to 224$\times$224 and we train both ResNet-18 and DenseNet-18 for 50 epochs. As for Kaggle Pneumonia Detection, we employ RetinaNet which uses ResNet-18 as backbone. The default image size is 512$\times$512 and the batch size is 4.

\section{Ablation Study}
In this part, we conduct experiments on ChestX-ray14. As we have mentioned above, we use 70k unlabeled images for C2L pretraining and then use the rest labeled training set for fine-tuning. We report averaged AUROC performance and results of eight class are also provided (results of all fourteen classes are provided in the supplementary material). Note that to save space, the ablation study of $\theta$ (cf. Equation~\ref{momentum}) is put in the supplementary material.

\begin{table}[t]
	\centering
	\caption{Ablation study of proposed batch mixup, feature mixup and mixed consistency loss $\text{loss}_M$. We use single \textbf{Mix.} to denote the traditional mixup method~\cite{zhang2017mixup}. \textbf{Bat.} and \textbf{Feat.} are abbreviations for Batch and Feature, respectively. We report the performance of fourteen categories in ChestX-ray14. The best results are in bold while the second best are underlined.}
	\label{mixup}
	\scalebox{1.0}{
	\begin{tabular}{c|c|c|c|c|c}
		& ImageNet &Mix. & Bat. Mix. & Bat. Mix. + $\text{loss}_M$ & Bat. + Feat. Mix. + $\text{loss}_M$\\
		\hline
		\hline
		Average & 74.4 & 74.7 & 75.3 & 75.6 & \textbf{76.3} \\
		\hline
		Atelectasis & 80.0 & \underline{81.4} & 80.1 & 80.9 & \textbf{81.9}\\
		Cardiomegaly &65.3& \underline{68.2} & \textbf{68.4} & \textbf{68.4} & 67.9\\
		Effusion &74.9 & 74.9 & 74.3& \underline{75.3}& \textbf{75.7}\\
		Infiltration &68.4& 67.1 & 66.9 & \underline{68.6}& \textbf{68.9}\\
	    Mass & 79.4& 79.6 & 80.1 & \underline{80.2} & \textbf{80.7}\\
		Nodule &82.2 & 79.4 & 79.2 & \underline{80.3} & \textbf{82.9}\\
		Pneumonia &72.1& 73.2 & 73.3 & \underline{73.7}& \textbf{74.6}\\
		Pneumothorax &77.7& 80.7 & 80.6 & \underline{81.7}& \textbf{82.4}\\
		Consolidation &69.6 & 70.9 & \textbf{71.6} & \underline{71.5}& 71.2\\
		Edema &76.4& 73.6 & \underline{76.1} & 74.9& \textbf{77.0}\\
		Emphysema & 64.9& 66.2 & \underline{68.2}& \underline{68.2}& \textbf{68.6}\\
		Fibrosis &69.9 & 71.5 & 71.1 & \underline{72.0}& \textbf{72.1}\\
		Pleural Thickening &79.5 & 81.3 & \underline{81.9} & \textbf{82.5}& 81.7\\
		Hernia & 82.1 &77.4 & \textbf{82.9} & 79.8& \underline{82.8}\\
	\end{tabular}}
\end{table}

\begin{table}[h]
	\centering
	\caption{Influence of augmentation strategies and length of $Q$. \textbf{Mix.} represents the proposed mixup method which includes batch and feature mixup with mixed consistency loss. Note that we employ RandCrop (random crop) for all experiments.}
	\label{aug}
	\scalebox{1.0}{
	\begin{tabular}{c|c|c|c|c|c|c|c|c|c|c}
		\multirow{2}{*}{RandCrop}& \multirow{2}{*}{Rotation} & \multirow{2}{*}{Jigsaw} & \multirow{2}{*}{Dropout} & \multirow{2}{*}{cutout} & \multirow{2}{*}{cutout + Mix.} & \multicolumn{4}{|c|}{Length of $Q$} & \multirow{2}{*}{Average} \\
		& & & & & & $2^{11}$ & $2^{13}$ & $2^{15}$ & $2^{17}$ & \\
		\hline\hline
		\checkmark & & & & & &\checkmark & & & & 74.1 \\ 
		\checkmark & & & & & & &\checkmark & & & 74.5 \\ 
		\checkmark & & & & & & & & &\checkmark & 74.4 \\
		\checkmark & & & & & & & & \checkmark & & 74.9 \\
		\checkmark & \checkmark & & & & & & & \checkmark & & 75.2 \\
		\checkmark & \checkmark & \checkmark & & & & & & \checkmark & & 75.0 \\
		\checkmark & \checkmark & & \checkmark & & & & & \checkmark & & 74.9 \\
		\checkmark & \checkmark & & & \checkmark & & & & \checkmark & & 75.4 \\
		\checkmark & \checkmark & & & & \checkmark & & & \checkmark & & \textbf{76.3}
	\end{tabular}}
\end{table}

We first report the ablation results of proposed mixup approaches in Table~\ref{mixup}. We can see that the proposed batch mixup can already outperform the original mixup method~\cite{zhang2016colorful} by 0.6 point on average performance. This is because using the same $\lambda$ for shared batches may help maintain the consistency between batches. After adding mixed consistency loss, we can improve the batch mixup method by about 0.3 point. Since the goal of C2L is to learn powerful feature representations, we further apply mixup to generated features. Somewhat surprisingly, we can find that the propose feature mixup can be well integrated with $\text{loss}_M$ and surpass batch mixup by approximate 1 point. In summary, the proposed mixup strategies can outperform the original mixup method by 1.6 points.

Another important component of C2L is the augmentation strategies. An appropriate augmentation method should reasonably increase the diversity of augmented batches. Such characteristic may help pretrained models to learn representations which are discriminative enough to distinguish different radiographs. In Table~\ref{aug}, we investigate the effects of widely adopted augmentation strategies. It is normal to find that random rotation and cutout can enhance the performance by 0.3 point while adding jigsaw and dropout may degrade performance. Moreover, we find that simply increasing the length of $Q$ can be harmful. We argue the reason is that a longer queue may contain more useless features and thus reduce the attention of other useful representations. 

\section{Fine-tuning C2L Pretrained Models}
In this section, we compare C2L pretrained models with Model Genesis, ImageNet pretraining and MoCo~\cite{he2019momentum}. For pretraining datasets, we merge the training sets of ChestX-ray14 and CheXpert with radiographs in MIMIC-CXR and MURA to generate an unlabeled database containing approximate 700k images. For network architectures, we deploy ResNet-18 and DenseNet-121, both of which are widely used networks.
\begin{table}[h]
	\centering
	\caption{Results on ChestX-ray14. We use \textbf{MG} to represent Model Genesis~\cite{zhou2019models} while \textbf{MoCo} is for Momentum Contrast~\cite{he2019momentum}.}
	\label{chestxray14}
	\scalebox{1.0}{
	\begin{tabular}{c|c|c|c|c|c|c|c|c}
		& \multicolumn{4}{{c|}}{ResNet-18} & \multicolumn{4}{{c}}{DenseNet-121}\\
		& MG & ImageNet & MoCo & C2L & MG & ImageNet & MoCo & C2L\\
		\hline
		\hline
		Average &80.9& 81.5&81.4 &\textbf{83.5} &82.4 &82.9 &83.0 & \textbf{84.4}\\
		\hline
		Atelectasis &79.2 & 80.1 &79.8 &\textbf{82.1} &80.7 &81.2 &81.7 &\textbf{82.7} \\
		Cardiomegaly &85.9 & 87.7 &87.5 &\textbf{89.7} &88.3 &88.5 &89.2 &\textbf{90.5} \\
		Effusion &85.7 &86.2 &87.0 &\textbf{88.2} &87.0 &86.7 &86.6 &\textbf{87.9} \\
		Infiltration &67.8 &68.9 &68.5 &\textbf{70.9} &68.9 &69.6 &70.2 &\textbf{70.9} \\
		Mass &81.9 &82.5 &83.0 &\textbf{84.5} &83.6 &84.4 &84.0 &\textbf{86.3} \\
		Nodule &75.4 &75.2 &75.5 &\textbf{77.2} &77.0 &78.1 &77.8 &\textbf{79.8} \\
		Pneumonia &74.0 &74.3 &74.5 &\textbf{76.3} &74.4 &75.1 &75.7 &\textbf{76.3} \\
		Pneumothorax &85.1 &85.8 &85.1 &\textbf{87.8} &87.0 &86.8 &86.5 &\textbf{88.4} \\
		Consolidation &78.3 &78.6 &77.9 &\textbf{80.6} &80.0 &79.3 &79.8 &\textbf{80.7} \\
		Edema &86.9 &87.4 &87.2 &\textbf{89.4} &87.7 &88.2 &88.6 &\textbf{89.4} \\
		Emphysema &89.7 &89.8 &90.0 &\textbf{91.8} &91.0 &91.6 &90.7 &\textbf{93.0} \\
		Fibrosis &80.8 &81.8 &80.5 &\textbf{83.8} &82.6 &83.0 &82.3 &\textbf{85.1} \\
		Pleural Thickening &76.1 &76.2 &76.4 &\textbf{78.2} &76.8 &77.2 &77.5 &\textbf{78.3} \\
		Hernia &86.4 &86.8 &86.3 &\textbf{88.8} &88.9 &92.1 &91.6 &\textbf{92.2} \\
	\end{tabular}}
\end{table}

\textbf{ChestX-ray14.} We report fine-tuned AUROC results on the validation set. Besides ImageNet pretraining, we also perform experiments using Model Genesis (MG)~\cite{zhou2019models} and recently proposed MoCo~\cite{he2019momentum}. In Table~\ref{chestxray14}, the proposed C2L method surpasses other approaches by a significant margin. In fact, although MG and MoCo are able to achieve comparable results with ImageNet pretraining, they cannot surpass ImageNet pretrained model significantly. However, C2L pretrained model outperforms ImageNet pretraining by 2 points on ResNet-18. On DenseNet-121, C2L achieves 84.4\% averaged AUROC which is  1.5 points higher than ImageNet pretraining.

\begin{table}[t]
	\centering
	\caption{Results on CheXpert. We report the performance on six classes.}
	\label{chexpert}
	\scalebox{0.9}{
	\begin{tabular}{c|c|c|c|c|c|c|c}
		Method&Model&Average&Atelectasis & Cardiomegaly & Consolidation & Edema & Pleural Effusion \\
		\hline\hline
		MG & ResNet-18 &86.7& 79.8 & 80.0 &91.5 &91.3&90.9\\
		ImageNet &ResNet-18 &87.0&80.3 &79.6 & 91.9 &91.7 & 91.5 \\
		MoCo & ResNet-18 &87.1 & 80.3 & 79.4 & 92.5 & 92.0 & 91.1\\
		C2L & ResNet-18 &\textbf{88.2} &\textbf{81.1} & \textbf{81.4} & \textbf{93.0} & \textbf{92.9} & \textbf{92.6} \\
		\hline
		MG & DenseNet-121 &87.5 &80.6&81.0&92.7&91.9&91.1\\
		ImageNet & DenseNet-121 &87.9 &81.5 &81.9&92.4&92.1&91.7\\
		MoCo & DenseNet-121 &87.4 &81.5 &80.8.&92.0&91.4&92.0\\
		C2L & DenseNet-121 &\textbf{89.3}&\textbf{83.3} & \textbf{83.0} & \textbf{93.6} & \textbf{92.7} & \textbf{93.8}\\
	\end{tabular}}
\end{table}

\textbf{CheXpert.} Similar to ChestX-ray14, we fine-tune the pretrained models using the training set. We can see that ImageNet pretraining performs better than MG while MoCo achieves comparable results. In contrast, C2L generates better pretrained models. On ResNet-18, C2L outperforms ImageNet pretraining by 1.2 points. When it comes to DenseNet-121, the performance gap becomes 1.4 point.

\begin{table}[h]
	\centering
	\caption{Results of mean average precision (mAP) under different thresholds of predicted scores on Kaggle Pneumonia dataset. For each threshold, we only report predictions whose confidence scores are higher than this threshold.}
	\label{rsna}
	\scalebox{0.9}{
	\begin{tabular}{C{1.5cm}|C{1cm}|C{1cm}|C{1cm}|C{1cm}|C{1cm}|C{1cm}|C{1cm}}
		&0.2&0.3&0.4&0.5&0.6&0.7&0.8 \\
		\hline\hline
		MG &12.5&16.0&18.5&20.7&21.4&21.1&20.4\\
		ImageNet &13.5 &17.1&19.9&21.4&22.2&22.1&21.1\\
		MoCo &13.0&17.4&19.9&21.3&22.4&21.9&21.2\\
		C2L &\textbf{14.8} &\textbf{18.4} &\textbf{21.3}&\textbf{22.4}&\textbf{23.9}&\textbf{23.1}&\textbf{22.5}\\
	\end{tabular}}
\end{table}

\textbf{Kaggle Pneumonia Detection.} We use ResNet-18 as the backbone of RetinaNet. In Table~\ref{rsna}, it is obvious that C2L outperforms ImageNet pretraining on all thresholds significantly, especially on large thresholds. As for MoCo and MG, MoCo is marginally better than ImageNet pretraining while MG performs slightly worse.

\section{Conclusion}\label{key}
We proposed a self-supervised pretraining method C2L (Comparing to Learn) to learn medical representations from unlabeled data. Our approach makes use of the relation between images as supervision signal and thus requires no extra manual labeling.

\section{Acknowledgment}
This work was funded by the Key Area Research and Development Program of Guangdong Province, China (No. 2018B010111001), National Key Research and Development Project (2018YFC2000702) and Science and Technology Program of Shenzhen, China (No. ZDSYS201802021814180).

%
%
%
%

	{\small
	\bibliographystyle{splncs04}
	\bibliography{egbib}
	}
\end{document}